\documentclass[10pt,twocolumn,letterpaper]{article}

\usepackage{iccv}
\usepackage{times}
\usepackage{epsfig}
\usepackage{graphicx}
\usepackage{amsmath}
\usepackage{amssymb}
\usepackage{graphicx,multirow,tabu}
\usepackage{booktabs}

\usepackage[pagebackref=true,breaklinks=true,letterpaper=true,colorlinks,bookmarks=false]{hyperref}
\usepackage{graphicx}
\usepackage{subcaption}
\makeatletter
\def\UrlAlphabet{%
      \do\a\do\b\do\c\do\d\do\e\do\f\do\g\do\h\do\i\do\j%
      \do\k\do\l\do\m\do\n\do\o\do\p\do\q\do\r\do\s\do\t%
      \do\u\do\v\do\w\do\x\do\y\do\z\do\A\do\B\do\C\do\D%
      \do\E\do\F\do\G\do\H\do\I\do\J\do\K\do\L\do\M\do\N%
      \do\O\do\P\do\Q\do\R\do\S\do\T\do\U\do\V\do\W\do\X%
      \do\Y\do\Z}
\def\UrlDigits{\do\1\do\2\do\3\do\4\do\5\do\6\do\7\do\8\do\9\do\0}
\g@addto@macro{\UrlBreaks}{\UrlOrds}
\g@addto@macro{\UrlBreaks}{\UrlAlphabet}
\g@addto@macro{\UrlBreaks}{\UrlDigits}
\makeatother

\usepackage{listings}
\usepackage{color}
\definecolor{gray}{rgb}{0.6,0.6,0.6}
\definecolor{codegreen}{rgb}{0,0.5,0}
\definecolor{codeblue}{rgb}{0.25,0.5,0.5}
\definecolor{codegray}{rgb}{0.6,0.6,0.6}
\iccvfinalcopy 

\def\iccvPaperID{3926} 

\ificcvfinal\fi
\begin{document}

\title{
    LIP: Local Importance-based Pooling
}

\author{
    Ziteng Gao \quad Limin Wang\thanks{Corresponding author.} \quad Gangshan Wu\\
State Key Laboratory for Novel Software Technology, Nanjing University, China\\
}

\maketitle
\ificcvfinal\thispagestyle{empty}\fi

\newcommand{\tabA}{
\begin{table}[]\small
    \centering
    \renewcommand\arraystretch{1.0}
    \setlength{\tabcolsep}{4pt}
    \begin{tabular}{lcccc}
        \toprule
    Method              & Top-1 & Top-5 & \#Params & FLOPs \\\midrule
    Strided convolution & 76.40 & 93.15 & 25.6M & 4.12G    \\
    Average pooling     & 76.96 & 93.35 & \textbf{22.8M} & \textbf{3.82G} \\
    DPP {\em \scriptsize (our baseline structure)} & 76.87 & 93.30 & \textbf{22.8M} & 3.83G\\
    \color{gray} DPP {\scriptsize {\em (original structure in }~\cite{DBLP:conf/cvpr/SaeedanWG018}\em )} & \color{gray} 77.22 & \color{gray} 93.64 & \color{gray} 25.6M & \color{gray} 6.59G \\
    \midrule
    LIP w Projection     & 77.49 & 93.86 & 24.7M & 4.78G \\
    LIP w Bottleneck-64  & 77.92 & 93.97 & 23.2M & 4.65G \\
    LIP w Bottleneck-128 & \textbf{78.19} & 93.96 & 23.9M & 5.33G \\
    LIP w Bottleneck-256 & 78.15 & \textbf{94.02} & 25.8M & 7.61G \\ \bottomrule
    \end{tabular}
    \vspace{-0.5em}
    \caption{ResNet-50 with different downsampling methods.
    }
    \label{tab:effective}
    \vspace{-1.0em}
\end{table}
}

\newcommand{\tabB}{
\begin{table}[]\small
    \centering
    \renewcommand\arraystretch{1.0}
    \setlength{\tabcolsep}{5pt}
\begin{tabular}{lccccc}
    \toprule
    \multirow{2}{*}{Layer}    & \multicolumn{5}{c}{Combination of LIP substitutions} \\
    \cmidrule(lr){2-6}
    & A & B & C & D & E
    \\
    \midrule
    Max Pooling  & \checkmark &            &            &           &   \\
    Res$_3$     & \checkmark & \checkmark &            &           &   \\
    Res$_4$     & \checkmark & \checkmark & \checkmark &           &   \\
    Res$_5$     & \checkmark & \checkmark & \checkmark & \checkmark&  \\
    Top-1    & 78.19 & 77.87 & 77.78 & 76.92 & 76.40 \\
    Top-5    & 93.96 & 93.94 & 93.81 & 93.37 & 93.15 \\
    \#Params & 23.9M & 23.8M & 23.7M & 23.9M & 25.6M \\
    FLOPs    & 5.33G & 4.87G & 4.26G & 4.11G & 4.12G
    \\\bottomrule
    \end{tabular}
    \vspace{-0.5em}
    \caption{Different LIP substitution locations. Combination A stands for the ResNet-50 with full 7 LIPs (LIP-ResNet w Bottleneck-128) and E stands for the vanilla ResNet.}
    \label{tab:ablation}
    \vspace{-1.0em}
\end{table}
}

\newcommand{\tabC}{
\begin{table}[]\small
    \centering
    \renewcommand\arraystretch{1.0}
    \setlength{\tabcolsep}{4pt}
\begin{tabular}{lcccc}
    \toprule
    \multirow{2}{*}{Layer}    & \multicolumn{4}{c}{Combination of layers in the top} \\
    \cmidrule(lr){2-5}
    & A & B & C & D
    \\
    \midrule
    Affined IN        & \checkmark & \checkmark &            &  \\
    Amplified sigmoid & \checkmark &            & \checkmark &  \\
    Top-1    & 78.19 & N/A & 77.81 & 77.89  \\
    Top-5    & 93.96 & N/A & 93.86 & 93.86  \\
    \bottomrule
    \end{tabular}
    \vspace{-0.5em}
    \caption{Different top layers on logit modules. Combination D is trained with average pooling within first 2000 iterations and then with LIP to avoid numerical overflow due to $\exp(\cdot)$ operation along with noisy gradient during early iterations. Combination B falls in training although we tried various ways to avoid numerical problems.}
    \label{tab:top}
    \vspace{-1.0em}
\end{table}
}

\newcommand{\tabD}{
\begin{table}[]\small
    \centering
    \renewcommand\arraystretch{1.0}
    \setlength{\tabcolsep}{3pt}
    \begin{tabular}{lccrr}
        \toprule
    Architecture        & Top-1 & Top-5 & \#Params & FLOPs \\
    \midrule
    ResNet-50          & 76.40 & 93.15 & 25.6M & 4.12G \\
    LIP-ResNet-50      & 78.19 & 93.96 & 23.9M & 5.33G \\
    ResNet-101          & 77.98 & 93.98 & 44.5M & 7.85G \\
    LIP-ResNet-101     & 79.33 & 94.60 & 42.9M & 9.06G \\
    ResNet-152$^{\rm *}$         & 78.49 & 94.22 & 60.2M & 11.58G\\
    \midrule
    DenseNet-BC-121       & 75.62 & 92.56 & 8.0M & 2.88G \\
    LIP-DenseNet-BC-121  & 76.64 & 93.16 & 8.7M & 4.13G \\
    \bottomrule
    \end{tabular}
    \vspace{-0.5em}
    \caption{ResNets and DenseNets with and without LIP. For ResNet-152, we adopt the result trained by a similar recipe\protect\footnotemark.}
    \label{tab:architect}
    \vspace{-1.0em}
\end{table}
\footnotetext{\url{https://github.com/tensorpack/tensorpack/tree/master/examples/ResNet}}
}

\newcommand{\tabE}{
\begin{table}[]\small
    \centering
    \renewcommand\arraystretch{1.0}
    \setlength{\tabcolsep}{4pt}
    \begin{tabular}{llllllll}
        \toprule
        Backbone & AP  & AP$_{50}$ & AP$_{75}$ & AP$_{s}$ & AP$_m$ & AP$_l$ \\
        \midrule
        \multicolumn{6}{l}{\emph{Faster R-CNN w FPN results}} \\
        ResNet-50      & 37.7 & 59.3 & 41.1 & 21.9 & 41.5 & 48.7 \\
        LIP-ResNet-50 & 39.2 & 61.2 & 42.5 & 24.0 & 43.1 & 50.3 \\
        ResNet-101     & 39.4 & 60.7 & 43.0 & 22.1 & 43.6 & 52.1 \\
        LIP-ResNet-101   & 41.7 & 63.6 & 45.6 & 25.2 & 45.8 & 54.0 \\
        ResNeXt-101 & 40.7 & 62.1 & 44.5 & 23.0 & 44.5 & 53.6 \\
        \midrule
        \multicolumn{6}{l}{\emph{RetinaNet results}} \\
        ResNet-50   & 36.6 & 56.6 & 38.9 & 19.6 & 40.3 & 48.9
        \\
        LIP-ResNet-50 & 38.0 & 58.8 & 40.5 & 22.6 & 41.5 & 49.9\\
        ResNet-101 & 38.1 & 58.1 & 40.6 & 20.2 & 41.8 & 50.8 \\
        \bottomrule
        \end{tabular}
    \vspace{-0.5em}
    \caption{Faster R-CNN with FPN and RetinaNet with different backbones results on COCO 2017 \texttt{val} set. ResNeXt-101 stands for ResNeXt-64x4d-101 backbone in \cite{DBLP:conf/cvpr/XieGDTH17}.}
    \label{tab:fpn}
    \vspace{-1.0em}
\end{table}
}

\newcommand{\tabF}{
\begin{table*}[] \small
    \centering
    \renewcommand\arraystretch{1.0}
    \setlength{\tabcolsep}{6pt}
    \begin{tabular}{llllllll}
        \toprule
        Detection Framework & Backbone & AP & AP$_{50}$ & AP$_{75}$ & AP$_s$ & AP$_m$ & AP$_l$ \\
        \midrule
        Faster R-CNN w FPN \cite{DBLP:conf/cvpr/LinDGHHB17} & ResNet-101 w FPN
        & 36.2 & 59.1 & 39.0 & 18.2 & 39.0 & 48.2 \\
        Mask R-CNN \cite{DBLP:conf/iccv/HeGDG17} & ResNet-101 w FPN
        & 38.2 & 60.3 & 41.7 & 20.1 & 41.1 & 50.2 \\
        SOD-MTGAN \cite{DBLP:conf/eccv/BaiZDG18} & ResNet-101 w FPN
        & 41.4 & 63.2 & 45.4 & \underline{24.7} & 44.2 & 52.6 \\
        Grid R-CNN \cite{DBLP:journals/corr/abs-1811-12030} & ResNet-101 w FPN
        & 41.5 & 60.9 & 44.5 & 23.3 & 44.9 & 53.1 \\
        DCR \cite{DBLP:conf/eccv/ChengWSFXH18} & ResNet-101-Deformable w FPN
        & 41.7 & 64.0 & 45.9 & 23.7 & 44.7 & 53.4 \\
        TridentNet \cite{DBLP:journals/corr/abs-1901-01892} & ResNet-101
        & 42.7 & 63.6 & \underline{46.5} & 23.9 & \underline{46.6} & \textbf{56.6} \\
        Cascade R-CNN \cite{DBLP:conf/cvpr/CaiV18} & ResNet-101 w FPN
        & \underline{42.8} & 62.1 & 46.3 & 23.7 & 45.5 & 55.2 \\
        \midrule
        Faster R-CNN w FPN \& LIP & LIP-ResNet-101 w FPN
        & 42.0 & \underline{64.3} & 45.8 & \underline{24.7} & 45.2 & 52.3 \\
        Faster R-CNN w FPN \& LIP & LIP-ResNet-101-MD w FPN
        & \textbf{43.9} & \textbf{65.7} & \textbf{48.1} & \textbf{25.4} & \textbf{46.7} & \underline{56.3} \\
        \bottomrule
        \end{tabular}
    \vspace{-0.5em}
    \caption{Results on COCO \texttt{test-dev} set.
    `Deformable' denotes deformable convolutions in \cite{DBLP:conf/iccv/DaiQXLZHW17}. `MD' denotes adding more deformable convolutions. The 1st and 2nd of each criterion are boldface and underlined respectively.}
    \label{tab:sota}
    \vspace{-1.5em}
\end{table*}
}

\newcommand{\figA}{
    \begin{figure*}[]
        \centering
        \begin{subfigure}{0.75\columnwidth}
            \centering
            \includegraphics[width=\columnwidth, trim=0 20 0 0,clip]{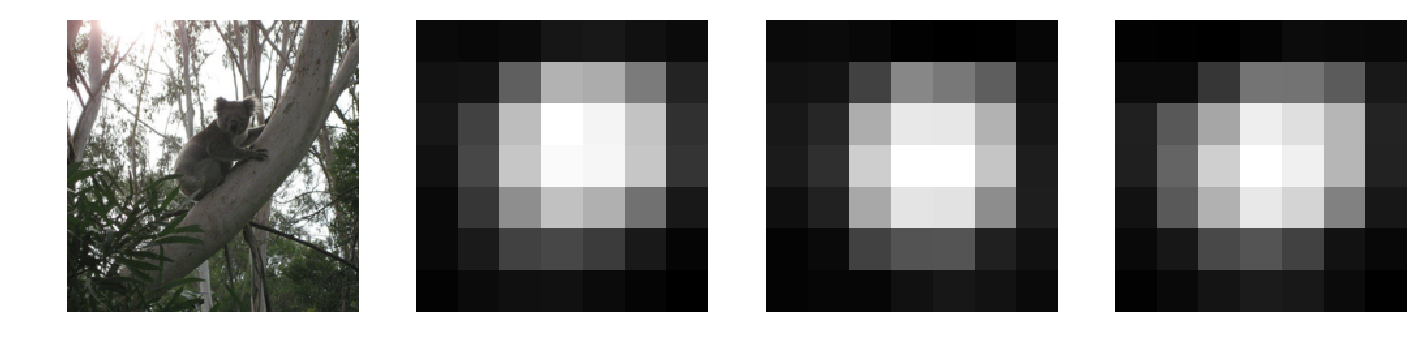} 
            \caption{}
        \end{subfigure}
        \begin{subfigure}{0.75\columnwidth}
            \centering
            \includegraphics[width=\columnwidth, trim=0 20 0 0,clip]{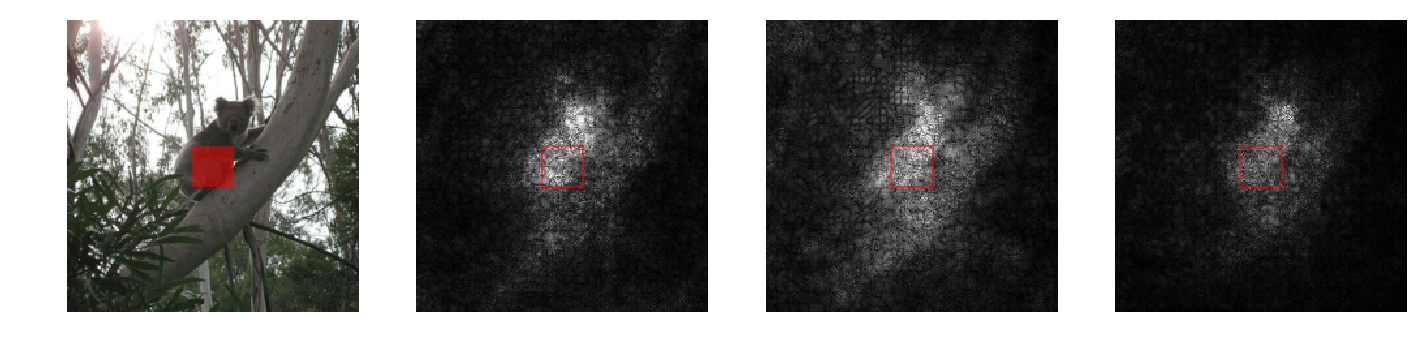}
            \caption{}
        \end{subfigure}

        \begin{subfigure}{0.75\columnwidth}
            \centering
            \includegraphics[width=\columnwidth, trim=0 20 0 0,clip]{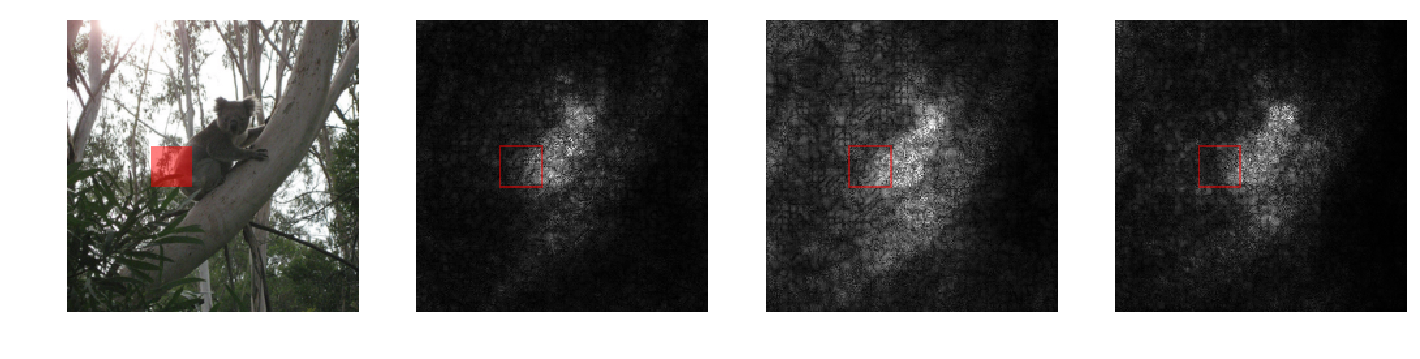}
            \caption{}
        \end{subfigure}
        \begin{subfigure}{0.75\columnwidth}
            \centering
            \includegraphics[width=\columnwidth, trim=0 20 0 0,clip]{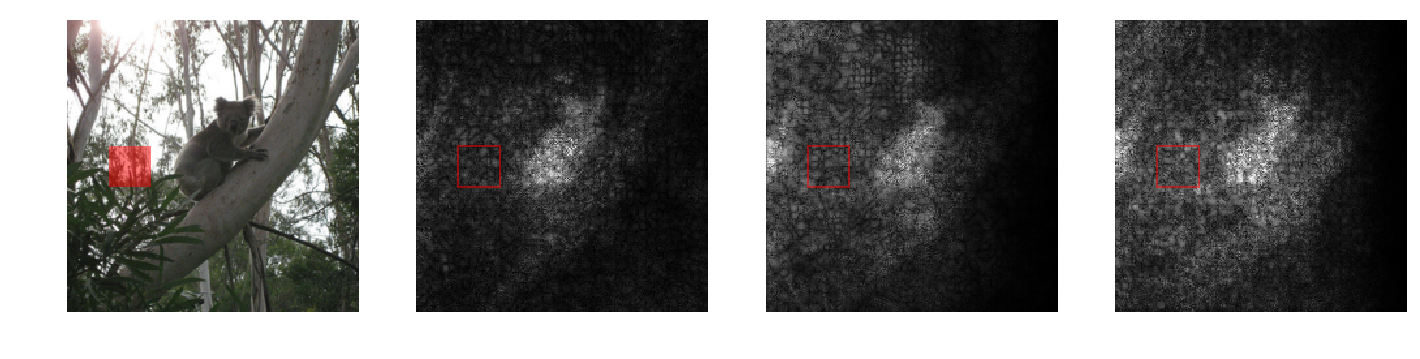}
            \caption{}\label{vis:last}
        \end{subfigure}
    \vspace{-0.8em}
    \caption{Visualizations of ResNets with different downsampling techniques. For each subfigure, there are an original image and results of ResNet-50 with LIP, average pooling and strided convolution from left to right. (a) denotes the class activation mappings (CAMs)~\cite{DBLP:conf/cvpr/ZhouKLOT16} for koala. (b)$\sim$(d) denote the effective receptive fields~\cite{DBLP:conf/nips/LuoLUZ16} in the image context, namely, backpropagated gradients from specific locations in CAMs (red in original images). Contrast of visualizations is lowered for human vision.
    }
    \label{fig:visualization}
    \vspace{-0.5em}
    \end{figure*}
}

\newcommand{\figB}{
    \begin{figure}[]
        \centering
        \includegraphics[width=0.8\columnwidth]{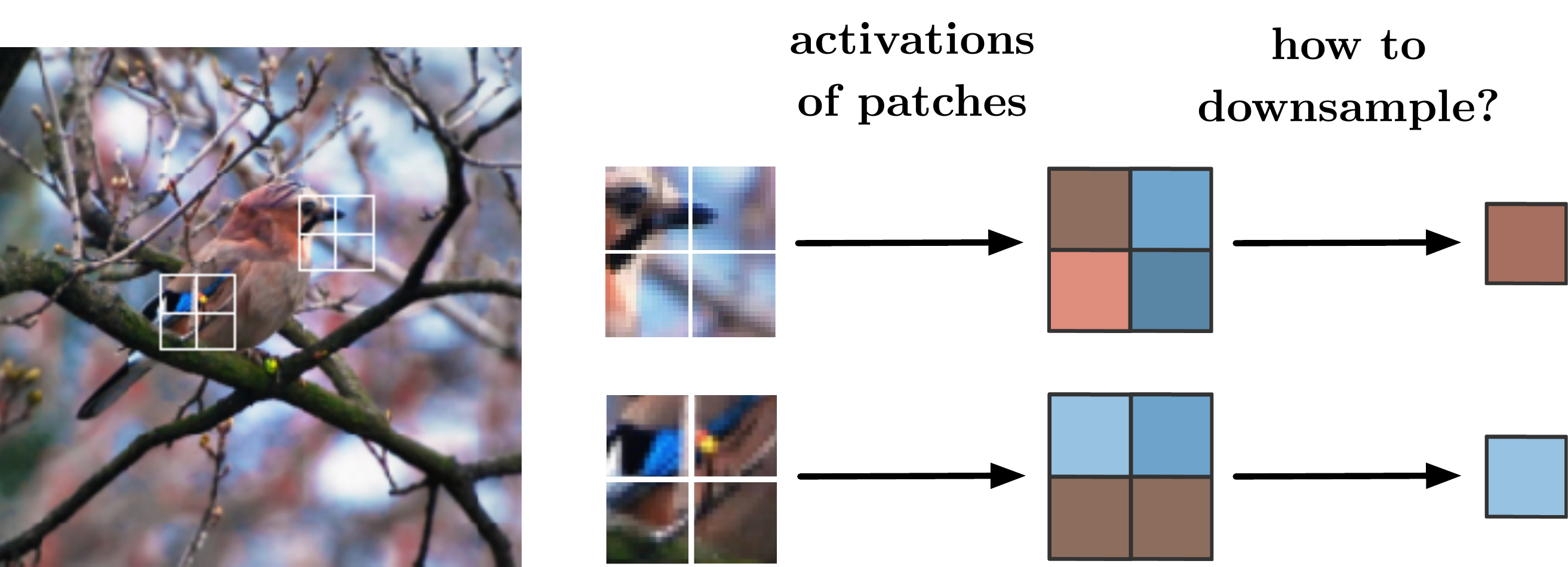}
    \vspace{-0.5em}
    \caption{Illustration of our motivation. (Left to right) the original image, some nearby patches, corresponding illustrative activations in the feature map to downsample, and last, the output activations that we need. Here, red-tone activations are caused by the foreground bird. Blue-tone activations are caused by the background clutter in the top patch or the representive blue feather in the bottom patch. We want to preserve the red-tone activations in the top patch window and the blue-tone activations in the bottom. The downsampling method should recognize discriminative features adaptively across sliding windows.}
    \label{fig:illustration}
    \vspace{-1.0em}
    \end{figure}
}

\newcommand{\figC}{
    \begin{figure*}[]
        \centering
        \begin{subfigure}{0.2\textwidth}
            \includegraphics[width=\textwidth]{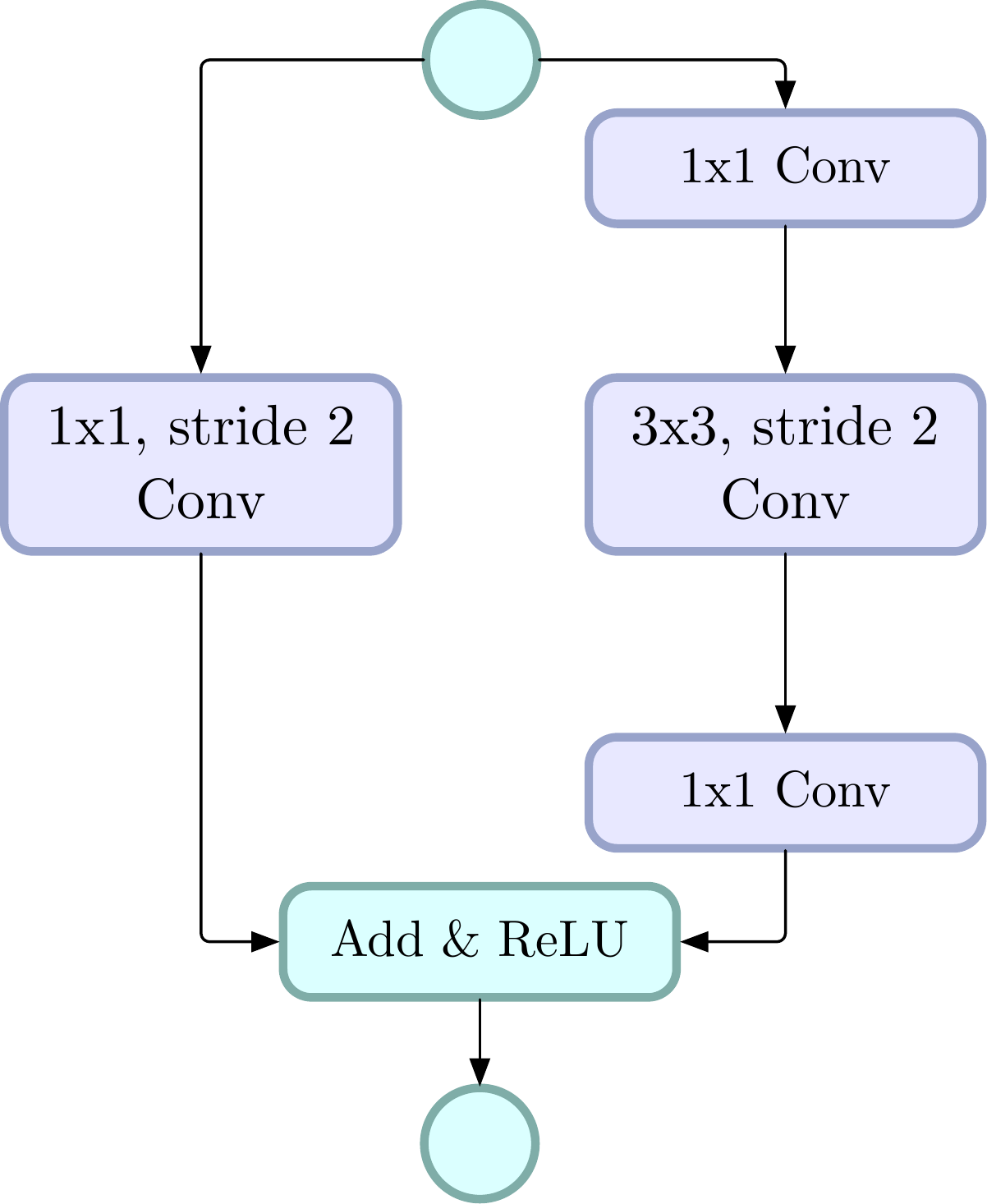}
            \caption{}
            \label{net:strided_conv}
            \vspace{-1em}
        \end{subfigure}
        \begin{subfigure}{0.2\textwidth}
            \includegraphics[width=\textwidth]{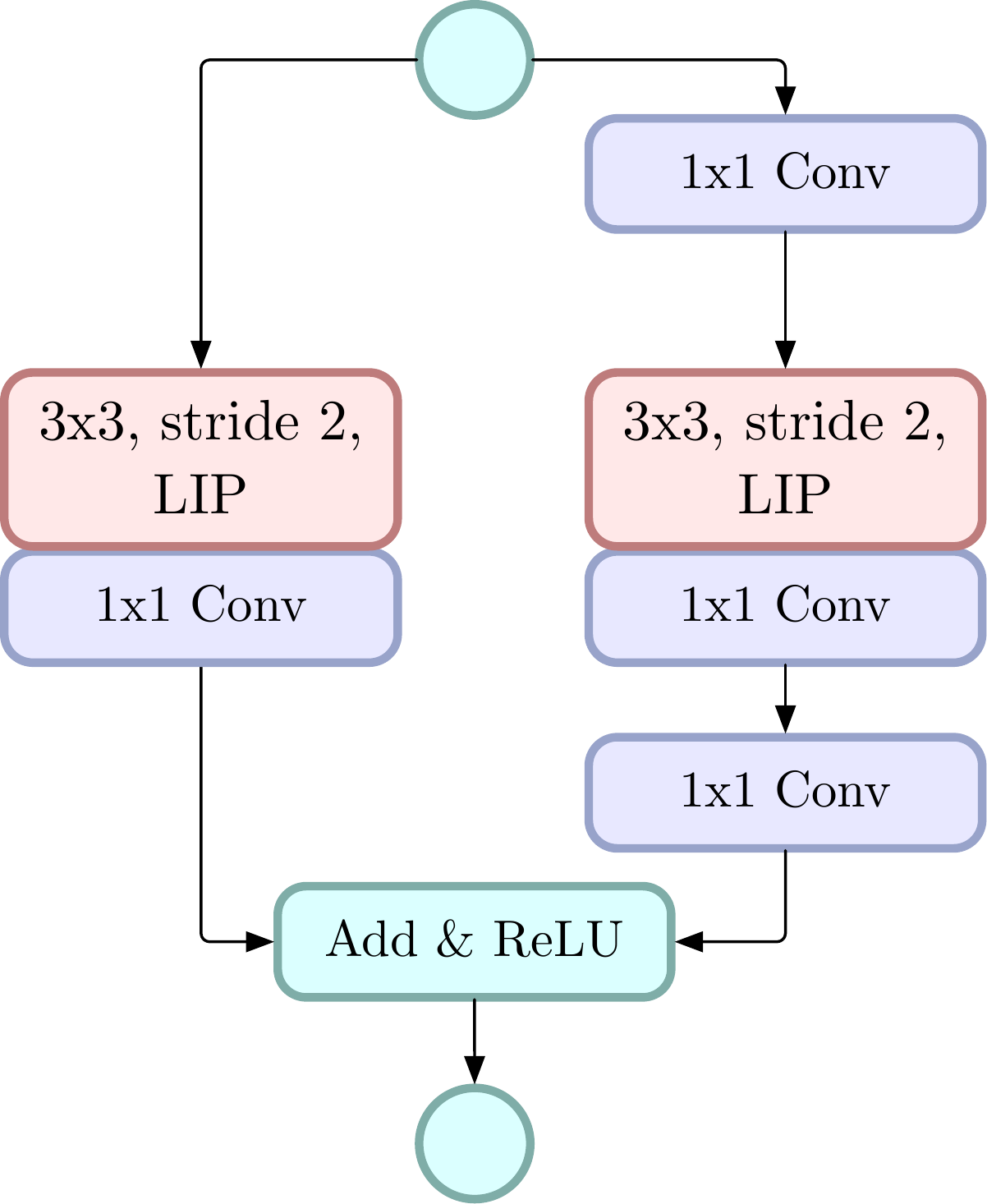}
            \caption{}
            \label{net:lip}
            \vspace{-1em}
        \end{subfigure}
        \begin{subfigure}{0.2\textwidth}
            \includegraphics[width=\textwidth]{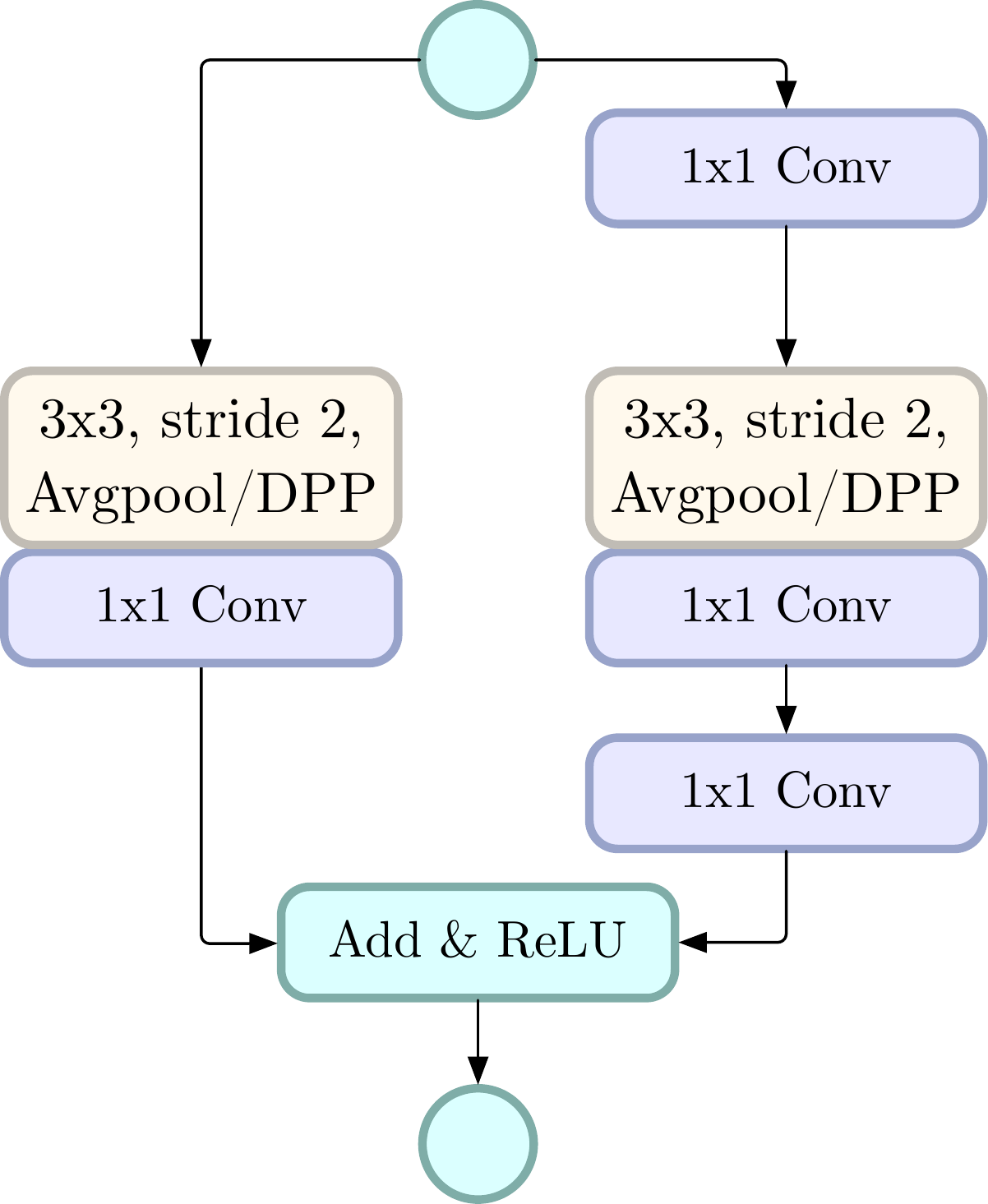}
            \caption{}
            \label{net:avgpool}
            \vspace{-1em}
        \end{subfigure}
        \begin{subfigure}{0.18\textwidth}
            \includegraphics[width=\textwidth]{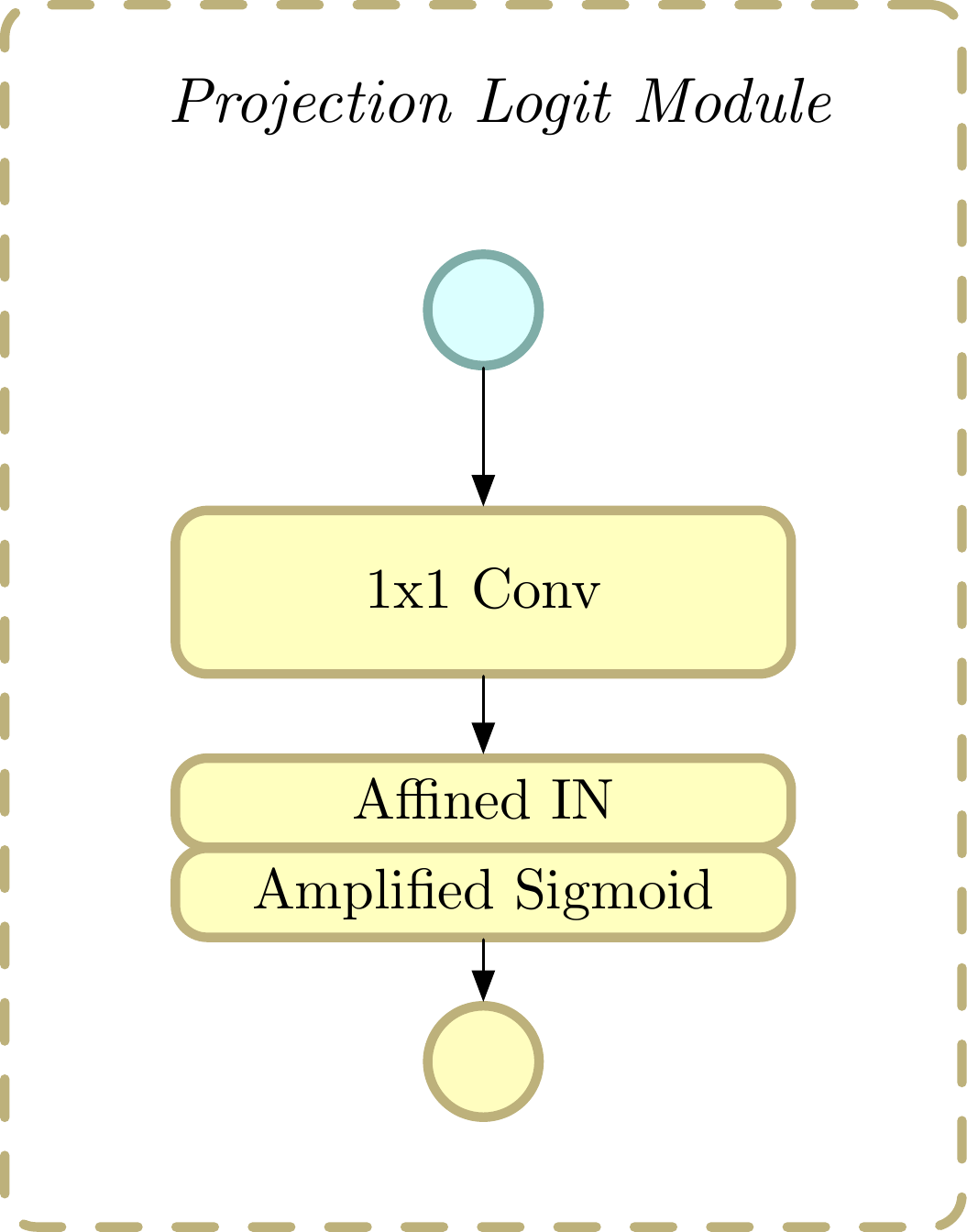}
            \caption{}
            \vspace{-1em}
            \label{module:projection}
        \end{subfigure}
        \begin{subfigure}{0.18\textwidth}
            \includegraphics[width=\textwidth]{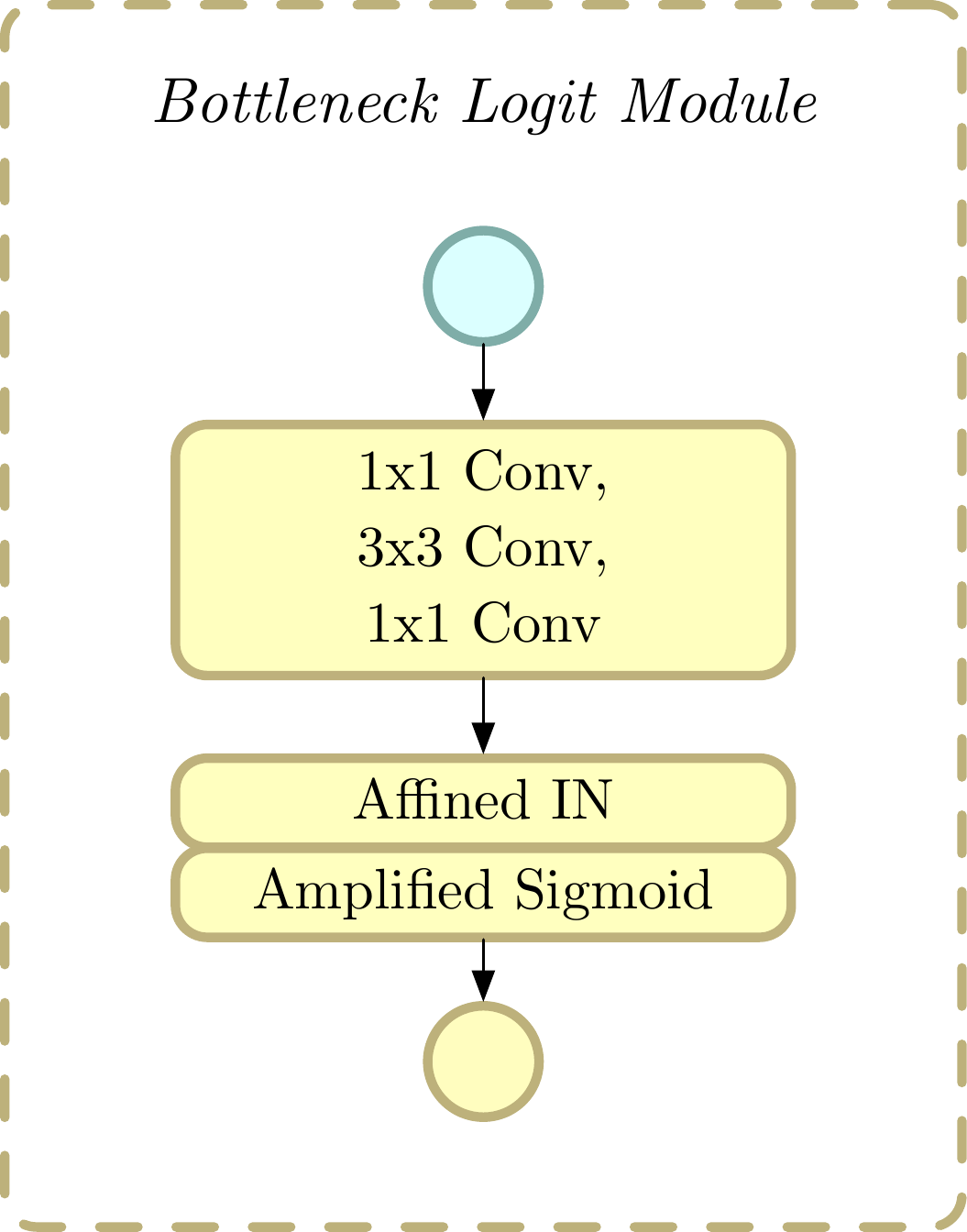}
            \caption{}
            \label{module:bottleneck}
            \vspace{-1em}
        \end{subfigure}
    \vspace{0.0em}
    \caption{Structures of ResNet building blocks for downsampling and logit modules.
    There are ResNet building blocks with strided convolutions~(a), LIP~(b), average pooling or DPP~(c).
    (d) and (e) show the projection and bottleneck logit module. The first two `Conv's in (e) mean convolutions and following affine Instance Normalization and ReLU function. }
    \label{fig:structure}
    \vspace{-1em}
    \end{figure*}

}

\newcommand{\figD}{
    \begin{figure*}[]
        \centering
        \includegraphics[width=0.75\textwidth]{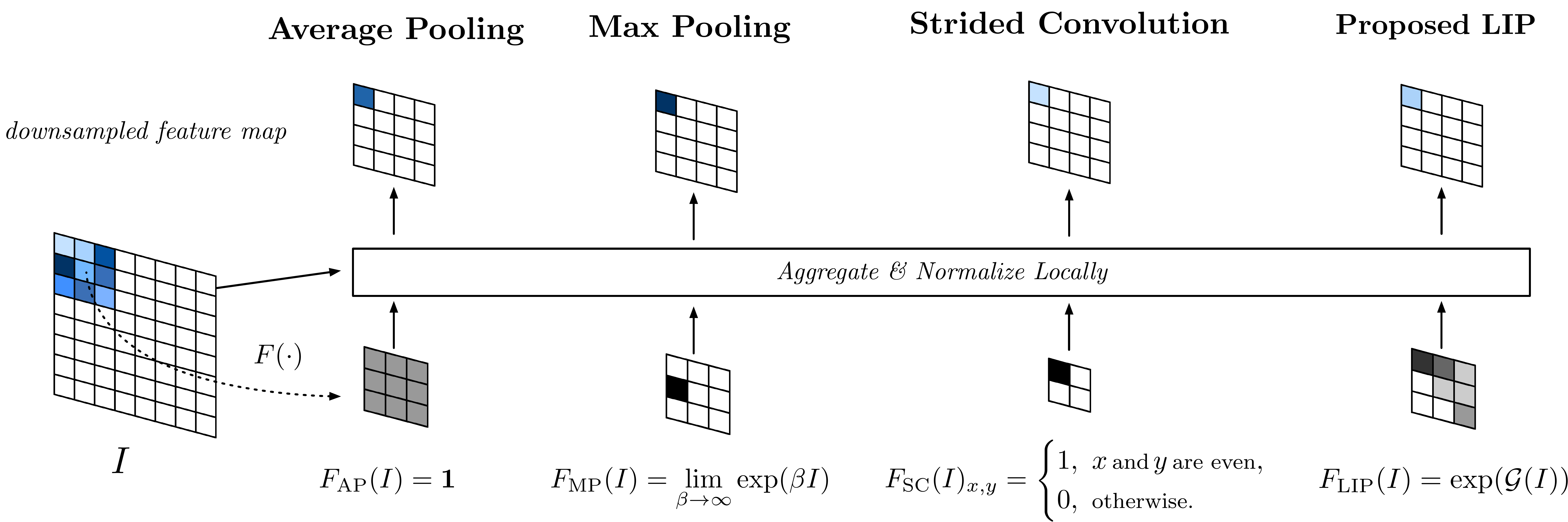}
    \vspace{-0.5em}
    \caption{
        Different downsampling methods viewed in the LAN framework. Activations in the input feature map are blue colored, darker meaning larger. Only activations and corresponding importance within current sliding window are shown. For strided convolution, the window size is equivalent to the stride, which is $2$ here.
        }
    \label{fig:princ}
    \vspace{-1.5em}
    \end{figure*}

}

\begin{abstract}
Spatial downsampling layers are favored in convolutional neural networks (CNNs) to downscale feature maps for larger receptive fields and less memory consumption. However, for discriminative tasks, there is a possibility that these layers lose the discriminative details due to improper pooling strategies, which could hinder the learning process and eventually result in suboptimal models. In this paper, we present a unified framework over the existing downsampling layers (e.g.,  average pooling, max pooling, and strided convolution) from a local importance view. In this framework, we analyze the issues of these widely-used pooling layers and figure out the criteria for designing an effective downsampling layer. According to this analysis, we propose a conceptually simple, general, and effective pooling layer based on local importance modeling, termed as {\em Local Importance-based Pooling} (LIP). LIP can automatically enhance discriminative features during the downsampling procedure by learning adaptive importance weights based on inputs. Experiment results show that LIP consistently yields notable gains with different depths and different architectures on ImageNet classification. In the challenging MS COCO dataset, detectors with our LIP-ResNets as backbones obtain a consistent improvement ($\ge 1.4\%$) over the vanilla ResNets, and especially achieve the current state-of-the-art performance in detecting small objects under the single-scale testing scheme.\footnote{Code is available at \url{https://github.com/sebgao/LIP}.}

\end{abstract}

\section{Introduction}
\figB
For discriminative tasks like image classification~\cite{DBLP:conf/cvpr/DengDSLL009} and object detection~\cite{DBLP:conf/eccv/LinMBHPRDZ14}, the modern architectures of convolutional neural networks (CNNs) mostly utilize spatial downsampling (pooling) layers to reduce the spatial size of feature maps in the hidden layers. Such pooling layers are for larger receptive fields and less memory consumption, especially in extremely deep networks~\cite{DBLP:journals/corr/SimonyanZ14a,DBLP:conf/cvpr/HeZRS16}. The widely-used max pooling, average pooling, and strided convolution use a sliding window whose stride is larger than $1$ and pool features by different strategies in each local window. But these layers might prevent discriminative details from being well preserved, which are crucial for recognition and detection task. This is especially undesirable for discriminative features of tiny objects, as such details might be diluted with clutter activations or even not be sampled by improper downsampling strategies.

In this paper, we aim to address these issues raised by the existing downsampling layers. To analyze their drawbacks, we present a unified framework from a local importance view. Under this new perspective, the existing pooling procedure could be seen as aggregating features with their local importance in each sliding window. To our best knowledge, we are the first to present a framework from the importance view for downsampling layers, which allows us to analyze and improve the pooling methods in a more principled way. As a result, we show that average and max pooling are both suboptimal due to the strong assumption or the invalid prior knowledge. Strided convolution adopts the improper interval sampling and also fails to model importance adaptively. To overcome their limitations, we present a new pooling method to learn importance weights automatically, coined as {\em Local Importance-based Pooling} (LIP).

Basically, we argue that not all nearby pixels contribute equally and some features are more discriminative than the others within a neighborhood in the downsampling procedure, as illustrated in Figure~\ref{fig:illustration}. Therefore, it is expected to explicitly model the local importance and build a metric measure over pixels within local neighborhoods. From this analysis, we propose the LIP to meet the requirement of an ideal pooling operation. Specifically, LIP proposes to learn the metric of importance by a subnetwork based on the input features automatically. In this sense, LIP is able to adaptively determine which features are more important to be kept through downsampling. For instance, LIP enables the network to preserve features of tiny targets while discarding false activations of the background clutter when recognizing or detecting small objects. Moreover, LIP is a more generic pooling method than the existing methods, in sense that it is capable of mimicking the behavior of average pooling, max pooling and detail-preserving pooling~\cite{DBLP:conf/cvpr/SaeedanWG018}.

Experiments show LIP outperforms baseline methods by a large margin on ImageNet~\cite{DBLP:conf/cvpr/DengDSLL009} with different architectures. We also evaluate our LIP backbones on the challenging COCO detection task~\cite{DBLP:conf/eccv/LinMBHPRDZ14}, where localizing small objects play an important role. The both one- and two-stage detectors with our LIP-ResNets as backbones obtain a consistent improvement over the vanilla ResNets, and in particular achieve a new  state-of-the-art performance in detecting small objects under the single-scale testing scheme.

\section{Related Work}
Downsampling layers as basic layers in CNNs were proposed with LeNet-5~\cite{lecun1998gradient} as a way to reduce spatial resolution by summing out values in a sliding window. Spatial downsampling procedures also exist in traditional methods. For example, HOG and SIFT~\cite{DBLP:conf/cvpr/DalalT05,DBLP:journals/ijcv/Lowe04} aggregated the gradient descriptors within each spatial neighborhood. Bag of words (BoW) based models also used intensive pooling in object recognition as to obtain more robust representations against translation and scale variance~\cite{DBLP:conf/iccv/SivicZ03,DBLP:conf/cvpr/LazebnikSP06}.

Modern CNNs utilize pooling layers to downscale feature maps mainly for larger receptive field and less memory consumption. VGG~\cite{DBLP:journals/corr/SimonyanZ14a}, Inception~\cite{DBLP:conf/cvpr/SzegedyLJSRAEVR15,DBLP:conf/icml/IoffeS15, DBLP:conf/cvpr/SzegedyVISW16} and DenseNet~\cite{DBLP:journals/corr/HuangLW16a} used average and max pooling as downsampling layers. ResNet~\cite{DBLP:conf/cvpr/HeZRS16} adopted convolutions whose stride is not $1$ to extract features at regular non-consecutive locations as downsampling layers.

Some pooling methods, including global average pooling~\cite{DBLP:journals/corr/LinCY13}, ROI pooling~\cite{DBLP:conf/iccv/Girshick15}, and ROI align~\cite{DBLP:conf/iccv/HeGDG17}, aim to downscale feature maps of arbitrary size to a fixed size and therefore enable the network to cooperate with inputs of different sizes. We do not discuss these methods as they are designed to specific architectures. Here, we only focus on pooling layers inside networks, that is, the ones that gradually downscale feature maps by a fixed ratio.

There are some analysis on pooling methods before the widespread application of CNNs. Boureau {\em et al.}~\cite{DBLP:conf/icml/BoureauPL10} analyzed average and max pooling in traditional methods, and proved that max pooling can preserve more discriminative features than average pooling in terms of probability. The work~\cite{DBLP:conf/iccv/XieZSYL15, DBLP:journals/sigpro/WangGLM17} showed that pooling can be without specific forms and learning to pool features is beneficial. Our work mainly follows this research line and our results further support these conclusions.

Recent work about pooling has focused on how to better downscale feature maps in CNNs through new pooling layers. Fractional pooling~\cite{DBLP:journals/corr/Graham14a} and S3pool~\cite{DBLP:conf/cvpr/ZhaiWKCLZF17} tried to improve the way to perform spatial transformation of pooling, which is not the focus of our paper. Mixed and hybrid pooling~\cite{DBLP:conf/rskt/YuWCW14,DBLP:conf/aistats/LeeGT16} used various combinations of max and average pooling to perform downscaling. $L_p$ pooling~\cite{DBLP:conf/pkdd/GulcehreCPB14} aggregated activations in the $L_p$ norm way, which can be viewed as a continuum between max and average pooling controlled by the learned $p$. These methods can unify max and average pooling and further improve the performance of networks. However, they can simply learn better pooling method based on average pooling and max pooling, or the combination of them, but fails to provide more insights about general donwsampling methods. Saeedan {\em et al.}~\cite{DBLP:conf/cvpr/SaeedanWG018} argued that details should be preserved and redundant features can be discarded by proposed detail-preserving pooling (DPP). The detail criterion of DPP is relatively hand-crafted by calculating the deviation from statistics of pixels in sliding windows, which is heuristic and may be not optimal.

In this paper, we analyze widely-used pooling layers based on a local importance view, which has not been investigated in previous work. Our proposed LIP, naturally arisen from this concept, outperforms hand-crafted pooling layers by a large margin.

Attention-based methods are recently popular in computer vision community~\cite{DBLP:conf/cvpr/0004GGH18, DBLP:journals/corr/abs-1904-05873}. Our LIP can be also seen as a local attention approach designed for pooling, of which attention weights are in the softmax form. LIP mainly differs from other attention methods in two important aspects for the better compatibility with downsampling procedure: (1) attention weights are produced by local convolutions in logit modules and then normalized locally; (2) LIPs do not adopt the key-query schemes in attention modeling for achieving better shift invariance.

\section{Local Importance Modeling}
In this section, we first present the framework for downsampling layers from local importance modeling view. We discuss some widely-used pooling layers in this framework. Next, we describe our proposed local importance-based pooling (LIP), which naturally arises from this analysis. Finally, we show how to equip popular architectures with LIP layers and then obtain LIP-ResNet and LIP-DenseNet.

\figD
\subsection{Framework and Analysis}\label{section:framework}
To analyze the existing downsampling methods and well motivate our LIP, we present a unified framework for downsampling layers from the view of local importance, named {\em Local Aggregation and Normalization} (LAN). Specifically, given the input feature map $I$, the kernel indice set $\Omega$ consisting of relative sampling locations $(\Delta x, \Delta y)$ in a sliding window, and the left-top location $(x, y)$ corresponding to the sliding window in the input feature map with regrad to the output location $(x', y')$, the LAN framework is formulated as:
\begin{equation}\label{align:framework}
    O_{x',y'} = \frac{
        \sum_{(\Delta x, \Delta y)\in \Omega}
        F(I)_{x+\Delta x, y+\Delta y}I_{x+\Delta x, y+\Delta y}
        }{
        \sum_{(\Delta x, \Delta y)\in \Omega}
        F(I)_{x+\Delta x, y+\Delta y}
    },
\end{equation}
where $F(I)$ is the importance map whose size is the same with $I$ and $F(I) \ge 0$ over space. The division $(x/x', y/y')$ stands for the stride factor, e.g., $x=2x', y=2y'$ for $2\times 2$ stride. We simply denote a stride $2\times 2$ as $2$ in this paper.
As the name of the framework implies, pooling in this view can be seen two steps: {\bf aggregate} features with the importance $F(I)$ and {\bf normalize} them by importance within local sliding windows.
This framework can be extended naturally to the multi-channel situation.

One can see pooling in this framework as weighted sum over each window where weights are locally normalized importance:
\begin{equation}
    \frac{F(I)_{x+\Delta x, y+\Delta y}}{\sum_{(\Delta x, \Delta y)\in \Omega}F(I)_{x+\Delta x, y+\Delta y}},
\end{equation}
for $I_{x+\Delta x, y+\Delta y}$, which we term simply as local importance. Therefore, local importance stands for weights of features within a sliding window. We can analyze which features in downsampling procedures are more important than others nearby by $F(I)$.

Our motivation is that since the feature pooling procedure is intrinsically lossy as it squeezes large input into small output, it is necessary to carefully consider which features to sample and how to aggregate them in a small sliding window as shown in Figure~\ref{fig:illustration}. Sampled features should be discriminative enough for the target tasks. The LAN framework provides a principled way to understand and improve these pooling methods by studying the corresponding importance function $F$. Next, we analyze some widely-used downsampling layers in this framework and figure out the requirement of an ideal pooling operation. Figure~\ref{fig:princ} shows some of these downsampling methods viewing in the framework.

\textbf{Average and max pooling.}
As discussed in~\cite{DBLP:conf/icml/BoureauPL10}, given $F(I)=\exp(\beta I)$, $\beta=0$ gives average pooling and $\beta\to\infty$ gives max pooling. Average pooling associates features with the same importance to all locations during aggregation in a small window, while max pooling put all attention on the largest activation within a neighborhood. We argue that both of them are suboptimal. Average pooling harms discriminative but small features and cause blurry downsampled features due to the strong assumption of the local equality of features. Max pooling as an improvement over average pooling on feature selection, however, assumes that the most discriminative feature should be of the maximum activation. This assumption mainly has two drawbacks. First, the prior knowledge that the maximum activation stands for the most discriminative detail, may not be always true. Second, the max operator over sliding windows hinders gradient-based optimization since in the backpropagation gradients are assigned only to the local maximums, as discussed in~\cite{DBLP:conf/cvpr/SaeedanWG018}. These sparse gradients would further enhance this inconsistence, in sense that discriminative activations will never become maximums unless current maximums are suppressed.

\textbf{Strided convolutions.}
Strided convolutions can be seen as dense convolutions whose stride is $1$, followed by spatial subsampling~\cite{DBLP:conf/icml/Zhang19}. This spatial subsampling can be interpreted as downsampling in our framework with
\begin{equation}
    F(I)_{x, y}=\begin{cases}
        1, \text{ if } x\text{ and }y\text{ are both multiples of } s, \\
        0, \text{ } \text{otherwise},
    \end{cases}
\end{equation}
where $I$ is densely convolved features and $s$ is both the stride factor and sliding window size.
From this perspective, the downsampling part of strided convolutions fails to model the importance in downsampling procedures adaptively. Moreover, it focuses only on one fixed location within each sliding window and discards the rest.
This fixed interval sampling scheme will limit shift invariance, as convolutional patterns are required to appear at specific and non-consecutive locations to activate. In this sense, minor shifts and distortions can lead to great changes in downsampled features and thus disturb the shift invariance of CNNs~\cite{DBLP:conf/icml/Zhang19}.
For the case of strided $1\times 1$ convolutions, it is even worse since the feature map are not fully utilized~\cite{DBLP:journals/corr/abs-1812-01187} and it will incur gradient checkerboard problem~\cite{DBLP:conf/cvpr/PalacioFHRBD18}.

\textbf{Detail-preserving pooling.}
Recent proposed detail-preserving pooling (DPP)~\cite{DBLP:conf/cvpr/SaeedanWG018} uses the detail criterion as importance function $F$, which is measured by the deviations of features from the activation statistics in sliding windows. DPP solves the problem of max pooling by designing more sophisticated importance function and ensuring the continuity for better gradient optimization. However, the assumption in DPP is heuristic and the more detailed feature might be the less discriminative ones. For example, the background clutter could be more detailed than a bird of solid color in foreground. Therefore, DPP might preserve the less discriminative details to outputs. Hand-crafted importance functions in max pooling and DPP incorporate the general prior knowledge into downsampling procedure, which might lead to the inconsistence with the final target of discriminative tasks.

\textbf{Requirements of ideal pooling.}
From the analysis above, we can figure out the requirement of an ideal pooling layer.
First, the downsampling procedure is expected to handle minor shifts and distortions as much as possible, and thus should avoid adopting the fixed interval sampling scheme, i.e., $F$ used by strided convolutions.
Second, the importance function $F$ should be selective to the discriminative features rather than manually designed based on prior knowledge, i.e., $F$ used in max pooling and DPP. This discriminativeness measure should be adaptive to different tasks and automatically determined by the final objective.

\subsection{Local Importance-based Pooling}\label{section:method}
\begin{figure}
    \centering
\begin{subfigure}{0.46\columnwidth}
    \centering
    \includegraphics[width=0.95\columnwidth]{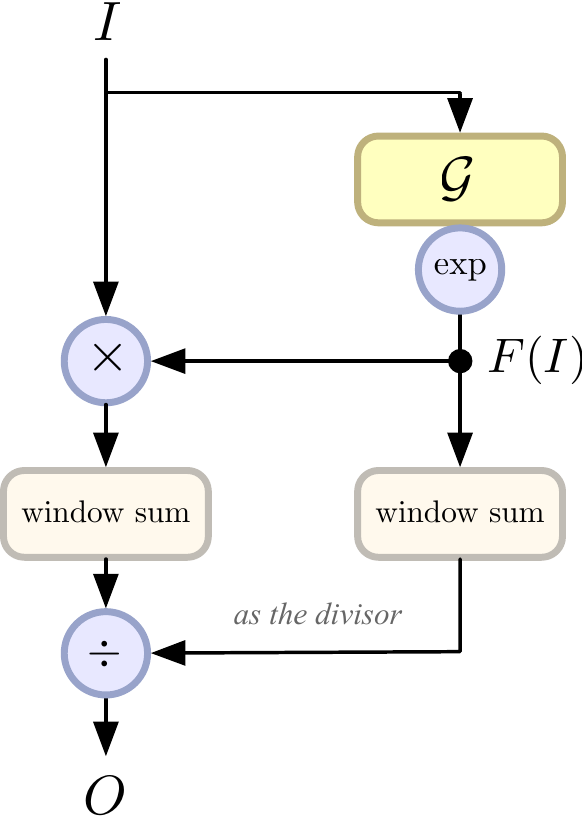}
    \caption{}
    \vspace{-1em}
\end{subfigure}
\begin{subfigure}{0.53\columnwidth}
    \vspace{2em}
    \begin{lstlisting}[language=python]
import torch
import torch.nn.functional as F

def lip2d(x, logit,
          kernel_size=3,
          stride=2,
          padding=1):
    weight = torch.exp(logit)
    return F.avg_pool2d(x*weight, kernel_size, stride, padding)/F.avg_pool2d(weight, kernel_size, stride, padding)
    \end{lstlisting}
    \vspace{0.7em}
    \caption{}
    \label{fig:code}
    \vspace{-1em}
\end{subfigure}
\caption{LIP operator and its PyTorch implementation. The logit module $\mathcal{G}$ is not shown in (b). The `window sum's in (a) mean locally summing within sliding windows.}
\label{fig:operator}
\vspace{-1.5em}
\end{figure}

To meet requirements of ideal pooling arisen from local importance view in the LAN framework, we propose local importance-based pooling (LIP).
By using a learnable network $\mathcal{G}$ in $F$, the importance function now is not limited in hand-crafted forms and able to learn the criterion for the discriminativeness of features. Also, we restrict the window size of LIP to be not less than stride to fully utilize the feature map and avoid the issue of fixed interval sampling scheme.
More specifically, the importance function in LIP is implemented by a tiny fully convolutional network (FCN)~\cite{DBLP:conf/cvpr/LongSD15}, which learns to produce the importance map based on inputs in an end-to-end manner. To make the importance weights non-negative and easy to optimize, we add $\exp(\cdot)$ operation on top of $\mathcal{G}$, that is:
\begin{equation}
    F(I)=\exp(\mathcal{G}(I)),
\end{equation}\label{align:function}
where $\mathcal{G}$ and $\mathcal{G}(I)$ are named the logit module and the logit respectively as $\mathcal{G}(I)$ is the logarithm of the importance.
In contrast to the hand-crafted form specified by prior knowledge in max pooling or DPP, the logit module $\mathcal{G}$ is able to learn a better and more compatible importance criterion for both the network and target task. More concretely, according to Equation (\ref{align:framework}), LIP is then written as:
\begin{equation}\label{align:pooling}
    O_{x',y'} = \frac{
        \sum_{(\Delta x, \Delta y)\in \Omega}
        {I_{x+\Delta x, y+\Delta y}\exp(\mathcal{G}(I))}_{x+\Delta x, y+\Delta y}
        }{
        \sum_{(\Delta x, \Delta y)\in \Omega}
        \exp(\mathcal{G}(I))_{x+\Delta x, y+\Delta y}
    }.
\end{equation}
With LIP, discriminative features can be automatically emphasized during downsampling procedure by learning a larger value of $\mathcal{G}(I)$ at the corresponding locations.
In the current implementation of LIP, the logit is calculated in a channel wise manner. Figure~\ref{fig:operator} shows the diagram and PyTorch implementation of LIP.

\textbf{Deformable modeling of LIP.}
At the macro level, learnable importance function $F$ of LIP enables the network to model deformation of objects by learning a good effective spatial allocation of features into downsampling with adaptive importance weights. Different from deformable convolutions~\cite{DBLP:conf/iccv/DaiQXLZHW17, DBLP:journals/corr/abs-1811-11168} to sample features by bilinear interpolation with adaptive offsets, LIP explicitly performs spatially dynamic feature selection based on inputs and thus has deformable receptive fields. Empirical evidence of the deformable capacity of LIP is shown and discussed in Section~\ref{sec:imagenet}.

\subsection{Exemplars: LIP-ResNet and LIP-DenseNet}\label{section:example}
ResNet~\cite{DBLP:conf/cvpr/HeZRS16} and DenseNet~\cite{DBLP:journals/corr/HuangLW16a} are typical architectures among modern CNNs.
ResNet mainly uses strided convolutions as downsampling layers except one max pooling in the bottom. DenseNet utilizes average pooling in transition blocks, and in the bottom a strided convolution layer and max pooling like ResNet to downscale feature maps.

\textbf{Architectures with LIP}.
We adopt the revised ResNet~\cite{revisedresnet} as our plain ResNet baseline, where residual branches employ $3\times 3$ kernel for strided convolutions, shown in Figure~\ref{net:strided_conv}.
To build LIP variants, we replace max pooling in the bottom and strided convolutions in downsampling blocks with LIP.
As discussed in Section~\ref{section:framework}, strided convolutions in ResNet could be replaced by a dense convolution and a following LIP. However, this substitution is computational intensive and memory inefficient. We instead first downscale features and then perform convolution. In this sense, we use a LIP and a following convolution to replace strided convolutions in residual and shortcut branches, as shown in Figure~\ref{net:lip}. To keep receptive fields the same and avoid the interval sampling problem, we set the window size of LIP to $3\times 3$ and the following convolution to $1\times 1$.
We leave the global average pooling in the top of ResNet unchanged. Total $7$ layers ($1$ for max pooling, $3\times 2$ for strided convolution) are replaced with LIP layers. We name this modified ResNet architecture as {\bf LIP-ResNet}.
For DenseNet, we replace $2\times 2$ average pooling layers in transition blocks and $3\times 3$ max pooling in the bottom by LIP layers of same configurations of window size. The global average pooling also remains unchanged like LIP-ResNet. Total $4$ layers ($1$ max pooling and $3$ average pooling) are replaced by LIP layers, and the resulted network is termed as {\bf LIP-DenseNet}.

\textbf{Design of logit modules.}
In the current implementation, we design two forms of logit modules for LIP layers, called the projection and the bottleneck form, respectively. Structures of logit modules are shown in Figure~\ref{module:projection} and~\ref{module:bottleneck}. In projection form, the logit module in LIP is simply composed of a $1\times 1$ convolution layer. The logit module of bottleneck form is like residual branches in bottleneck blocks~\cite{DBLP:conf/cvpr/HeZRS16}, which aims to capture spatial information in an efficient way. This form is denoted as Bottleneck-$x$, where $x$ is number of channels in the the input and output of $3\times 3$ convolution. To further reduce computational complexity of bottleneck logit modules in LIP-ResNet, the first $1\times 1$ convolution and $3\times 3$ convolution are shared between the residual and shortcut branches in a building block. The input of logit modules here is changed to the feature map fed into the building block, i.e., the top cyan circle in Figure~\ref{net:lip}, instead of the feature map to downsample. Bottleneck-$x$ logit module in LIP substitution for replacing max pooling in ResNet and DenseNet is simply a $3\times 3$ convolution.

For more effective modeling and stable training, we apply affine instance normalization~\cite{DBLP:journals/corr/UlyanovVL16} as spatial normalization and sigmoid function with a fixed amplification coefficient on the top of each logit module. Affine instance normalization make activations on each channel of each feature map follow normal distribution and then rescale it by learnable affine parameters.
The spatial normalization and rescale operation aim to help learn extreme cases such as max pooling.
The sigmoid function is used here to maintain numerical stability and the fixed amplification coefficient provides large enough range for logits, which is set to $12$ throughout our experiments.

\section{Experiments}
To validate the effectiveness of our LIP, we carry out experiments on the ImageNet 1K classification task~\cite{DBLP:conf/cvpr/DengDSLL009} and the MS COCO detection task~\cite{DBLP:conf/eccv/LinMBHPRDZ14}.

\subsection{ImageNet Classification Experiment Setup}
ImageNet 1K classification task~\cite{DBLP:conf/cvpr/DengDSLL009} requires the methods to cope with high-resolution images to capture discriminative details. We use (LIP-)ResNet and (LIP-)DenseNet for our experiments on the ImageNet classification task. For (LIP-)ResNet training, we use 8 GPUs and a mini-batch of 256 inputs, 32 images per GPU. For (LIP-)DenseNet training, we use 4 GPUs and a mini-batch of 256, 64 images per GPU. Our training procedure is generally following the recipe~\cite{DBLP:journals/corr/GoyalDGNWKTJH17} with two minor modifications. One is that we use SGD optimizer to update parameters with the vanilla momentum rather than Nesterov one. The other is that weight decay of $10^{-4}$ is applied to all learnable parameters including those of Batch Normalization.
All LIP layers are initialized to behave like average pooling by initializing parameters of the last convolution in logit modules to $0$. All results are reported on the validation set with single-crop testing.

\tabA
\subsection{Results on ImageNet and Analysis}\label{sec:imagenet}

\figC
\textbf{Study on LIPs and different logit modules.}
To compare with other pooling methods, we replace all LIP layers in LIP-ResNet by other pooling layers, i.e., average pooling or DPP, and keep the same configuration of window size and stride for fair comparison. The building blocks of these baselines are shown in Figure~\ref{net:avgpool}. Note that these baselines eliminate other factors including receptive fields and non-linearities to to be more consistent with LIP-ResNet. In this study, we resort to the ResNet-50 to perform comparison between different pooling layers.

\tabC
The results are reported in Table~\ref{tab:effective}. {\em First}, the ResNet-50 baseline with average pooling reduces both parameters and FLOPs, but still improves performance over the vanilla ResNet by around 0.5\% in top-1 accuracy. This result may be ascribed to the fixed interval sampling issue in strided convolutions, and a similar result was found in~\cite{DBLP:journals/corr/abs-1812-01187}.
{\em Second}, for our downsampling method, LIP with the simplest projection logit modules gains a noticeable improvement over these baselines ($>0.5\%$ in top-1). This shows that the importance simply learned from the projection logit module is beneficial for downsampling procedure.
{\em Third}, with a more powerful logit module Bottleneck-64, LIP-ResNet further improves accuracy over the projection one with fewer parameters and less computational cost. This demonstrates that spatial information is helpful for designing a better logit module.
The performance would saturate when we stretch the bottleneck logit module wider, and the Bottleneck-128 is a good trade-off between computational complexity and recognition performance, improving by 1.79\% in top-1 and 0.81\% in top-5 over the plain network. We adopt LIP with the Bottleneck-128 logit module as our default choice in the remaining experiments. 
{\em Finally}, we test the effectiveness of instance normalization and amplified sigmoid function. Results are shown in Table~\ref{tab:top}. The combination of them improves accuracy by enabling LIP to approximate extreme cases such as max pooling stably.

\textbf{LIP layers at various locations.}
\tabB
Table~\ref{tab:ablation} shows the results by placing different numbers of LIPs at different locations. We can find more LIPs generally contributes to better result but LIPs at different locations may not improve performance equally. LIP as the max pooling substitution only improves the top-1 accuracy significantly. We suspect that a single convolution as the logit module at this layer fails to encode enough semantic information to provide powerful logits into LIP. Another possible reason is that high-resolution details may help fine-grained classification but not benefit coarse-grained one. We can also find that the LIP at Res$_4$ is the most effective one. This might be due to the fact that the feature at this layer contains more semantics and the feature map size is still relatively large for downscaling. For practical applications, we recommend the usage of Combination C in Table~\ref{tab:ablation} due to less parameters and only 3\% extra FLOPs compared to the vanilla ResNet. But our default choice for the remaining experiments is the full LIP model, i.e., Combination A.

\textbf{Different network depth and architectures.}
We also evaluate LIP-ResNet and LIP-DenseNet with the deeper network, and the result is summarized in Table~\ref{tab:architect}. We find that LIP-ResNet-50 performs comparably to the vanilla ResNet-101 with only about half parameters and less FLOPs. LIP-ResNet-101 surpasses the vanilla ResNet-152 in both top-1 and top-5 accuracy by a notable margin (0.84\% and 0.38\%). For DenseNet and LIP-DenseNet, the result is also favorable, demonstrating the effectiveness of our method across different network architectures.
\tabD

\textbf{Visualizations.}
\figA
As discussed in Section~\ref{section:method}, LIP enables the network to have capacity of deformable modeling. To show this, we perform some visualizations of LIP layers. We first compute class activation mappings (CAMs)~\cite{DBLP:conf/cvpr/ZhouKLOT16} of ResNet-50 models with LIP, average pooling, and strided convolution. Next, we backpropagate activation of specific locations in CAMs to get gradient maps, which are called effective receptive fields~\cite{DBLP:conf/nips/LuoLUZ16} of specific locations in the original image context. Results are shown in Figure~\ref{fig:visualization}. The CAMs are similar but the gradient maps differ much among three downsampling approaches.
The effective receptive field of the model with LIP layers is more compact and mainly focuses on the foreground even when the backpropagated location moves out of the foreground (i.e., Figure~\ref{vis:last}).
Average pooling and strided convolution ones, however, are interfered more by the background clutter when backpropagating the activation out of the foreground.
This comparison shows the deformable modeling capacity of LIP layers. Compared with the average pooling and strided convolution, the clutter and background without discriminative features contribute much less to final recognition results in LIP-ResNet.

\subsection{MS COCO Detection Experiment Setup}
After verifying the effectiveness of LIP on image classification, we now focus on the more challenging detection task. 
There exists the problem of invisibility of tiny objects in most CNN architectures for detection~\cite{DBLP:journals/corr/abs-1804-06215}. This issue is mainly caused by losing discriminative information of small objects during improper downsampling procedure, which is suitable to justify the design of our LIP.

MS COCO detection~\cite{DBLP:conf/eccv/LinMBHPRDZ14} is a challenging task where the object scale variation is very large and detecting small objects plays a crucial role in final detection performance~\cite{DBLP:conf/cvpr/SinghD18, DBLP:conf/nips/SinghND18}.
We adopt mmdetection codebase~\cite{mmdetection2018} for our experiments. Our training strictly follows the default configuration of mmdetection, which includes setting shorter size of the image to 800, using standard horizontal flipping augmentation and ROI Align~\cite{DBLP:conf/iccv/HeGDG17}. In this experiment, we train two detection frameworks: Faster R-CNN with FPN~\cite{DBLP:conf/cvpr/LinDGHHB17} and RetinaNet~\cite{DBLP:conf/iccv/LinGGHD17}, on the COCO 2017 \texttt{train} set with the pre-trained backbone networks in Section~\ref{sec:imagenet}. We adopt the typical $2\times$ training time scheme for all COCO experiments. The baseline results are reported by evaluating the released detectors in mmdetection model zoo~\footnote{Evaluated when this paper was submitted and some baseline results are slightly higher than the officially reported ones in~\cite{mmdetection2018}.}. Detection performance is reported with single-scale testing.

\subsection{Results on MS COCO and Analysis}
\tabE
\tabF

The results of different backbones with Faster R-CNN and FPN are shown in Table~\ref{tab:fpn}. LIP-ResNet-50 and LIP-ResNet-101 backbones with Faster R-CNN yield 1.5\% and 2.3\% gain in AP over baselines, showing the effectiveness of our LIP at capturing discriminative features for detection branch. Their improvement gap may be ascribed to the fact that the deeper backbone provides more semantic features to produce better logits for LIP downsampling. 
For small object detection, the deeper vanilla ResNet only results in 0.2\% gain in AP$_s$, while LIP-ResNet-101 is better than LIP-ResNet-50 by 1.2\% in AP$_s$. The improvement of LIP-ResNets in AP$_s$ over the vanilla ResNets (2.1\% and 3.1\%) is also notable. These results show that the LIP layers are able to better preserve discriminative features of tiny objects. The results with the single-stage RetinaNet also validate the effectiveness of the LIP layer.

To compare with the state-of-the-art detectors, we train the deformable backbone (following the placement of more deformable convolutions in~\cite{DBLP:journals/corr/abs-1811-11168}, but without modulation and feature mimicking) with LIP in Faster R-CNN and FPN framework. The results are shown in Table~\ref{tab:sota}. The detectors with LIP-ResNet-101 are comparable to the state-of-the-art methods by simply using a standard detection pipeline without any specific design. The LIP-ResNet-101-MD backbone can further boosts AP to 43.9\% and AP$_s$ to 25.4\%, yielding a new state-of-the-art performance in detecting small objects under the single-scale testing scheme.

\section{Conclusion and Future Work}
In this paper, we stress spatial importance modeling in pooling procedures. We have presented the {\em Local Aggregation and Normalization} (LAN) framework based on local importance to analyze the widely-used pooling layers. Under the framework, we figure out these layers might keep out discriminative features due to using improper downsampling importance maps. Based on this analysis, we have proposed the {\em Local Importance-based Pooling} (LIP), a conceptually simple, general, and effective donwsampling method, with a goal of learning adaptive and discriminative importance maps to aggregate features for downsampling. Networks with LIPs are able to better preserve the discriminative details, especially those of tiny objects. Experiments on the ImageNet classification task indicate that LIP can capture rich details for holistic image recognition. On the COCO detection task, LIPs enable both one- and two-stage detection frameworks to yield better performance, especially that on small objects. Moreover, detectors with LIP-ResNet backbones reach a new state-of-the-art performance in detecting small objects by simply using a standard detection framework. 

In the future, we plan to study more aspects of implementation of LIP, such as logit module design, adaptive pooling size exploration and so on. Meanwhile, we will verify the effectiveness of LIP to more tasks, e.g., pose estimation and image segmentation.

\section*{Acknowledgments}
This work is supported by the National Science Foundation of China (No. 61921006, No. 61321491), and Collaborative Innovation Center of Novel Software Technology and Industrialization. The first author would like to thank Nan Wei and Qinshan Zeng for their comments and support.
\clearpage
{\small
\bibliographystyle{ieee_fullname}
\bibliography{camera_ready}
}

\end{document}